\documentclass[final]{l4dc2025}

\title[Game-Theoretic Motion Planning and Control in Autonomous Racing]{\hspace{-.8mm}$\alpha$-RACER: Real-Time Algorithm for Game-Theoretic Motion Planning and Control in Autonomous Racing using \(\alpha\)-Potential Function}
\usepackage{times}
\usepackage{macros}
\usepackage{longtable}
\author{%
 \Name{Dvij Kalaria}\thanks{Equal Contribution.} \Email{dvijk@berkeley.edu}\\
 \addr EECS, UC Berkeley
 \AND
 \Name{Chinmay Maheshwari\(^*\)} \Email{chinmay\_maheshwari@berkeley.edu}\\
 \addr EECS, UC Berkeley%
 \AND 
 \Name{Shankar Sastry} \Email{sastry@coe.berkeley.edu}\\
 \addr EECS, UC Berkeley
}

\begin{document}

\maketitle

\begin{abstract}
Autonomous racing extends beyond the challenge of controlling a racecar at its physical limits. Similar to human drivers, autonomous vehicles must also employ strategic maneuvers, such as overtaking and blocking, to gain an advantage over competitors.
While modern control algorithms can achieve human-level performance in single-car scenarios, research on real-time algorithms for multi-car autonomous racing remains limited.
To bridge this gap, we develop a game-theoretic modeling framework that incorporates the competitive aspects of autonomous racing, such as overtaking and blocking, through a novel policy parametrization, while operating the car at its limit. We propose an algorithmic approach to compute an approximate Nash equilibrium strategy for our game model, representing the optimal strategy for any autonomous vehicle in a competitive environment.
Our approach leverages a recently introduced framework of dynamic \(\alpha\)-potential functions, enabling the real-time computation.
Our approach comprises two phases: offline and online. During the offline phase, we use simulated racing data to learn an \(\alpha\)-potential function that approximates utility changes for agents. This function facilitates the online computation of approximate Nash equilibria by maximizing its value. We evaluate our method in a head-to-head 3-car racing scenario, demonstrating superior performance over several existing baselines.

\hspace{-7mm} \textbf{Website and Supplementary: } \href{https://sastry-group.github.io/alpha-RACER/}{https://sastry-group.github.io/alpha-RACER/} 
\end{abstract}


\section{Introduction}
Autonomous racing is a challenging task in autonomous vehicle development, requiring efficient planning, reasoning, and action in high-speed, dynamic, and constrained environments—key for addressing edge cases in broader autonomous driving. Recent advances, such as deep reinforcement learning (RL), have enabled vehicles to outperform human drivers \citep{wurman2022outracing, kaufmann2023champion}. However, challenges remain in optimizing strategies against other autonomous agents and in reducing the extensive training times required for achieving competitive performance.

A key challenge in multi-agent autonomous racing is developing real-time competitive strategies that consider the presence of other autonomous vehicles while maintaining high speeds \citep{betz2022autonomous}. Algorithms must balance lap-time optimization with aggressive driving, collision avoidance, and dynamic responses to competitors. While existing research on single-agent racing focuses on computing race lines based on vehicle dynamics and track constraints \citep{rosolia2017autonomous, balaji2020deepracer, song2021autonomous, herman2021learn, liniger2015optimization, kabzan2020amz}, these approaches cannot accommodate complexities of multi-agent settings, where interactions such as blocking and overtaking become crucial. To address these challenges, multi-agent racing requires strategies that account for the interdependent behaviors of agents, with frameworks like Nash equilibrium offering a way to anticipate and adapt to competitors' actions in a competitive environment.

Computation of Nash equilibrium is generally computationally challenging. Several studies have investigated its computation in autonomous racing, but many limitations persist. Many works use kinematic vehicle models \citep{notomista2020enhancing, wang2019bgame, wang2021game, liniger2019noncooperative, jia2023rapid, williams2017autonomous, sinha2020formulazero, schwarting2021stochastic}, which simplify vehicle dynamics but fail to capture the nonlinear tire forces that are critical for high-speed racing. Others rely on open-loop control via receding horizon techniques \citep{buyval2017deriving, jung2021game, wang2019bgame, wang2021game, liniger2019noncooperative, jia2023rapid, schwarting2021stochastic, zhu2023sequential, zhu2024sequential}, focusing on finite-horizon planning at the expense of long-term strategies. Additionally, many methods assume two-player zero-sum game \citep{wang2019bgame, thakkar2022hierarchical, kalaria2023towards,zheng2022game} or two-team competition \citep{thakkar2024hierarchical}, which are insufficient for races involving more than two agents. Alternative approaches, such as Stackelberg games \citep{notomista2020enhancing, liniger2019noncooperative}, iterative linear quadratic games \citep{schwarting2021stochastic}, and local Nash equilibrium \citep{zhu2023sequential, zhu2024sequential}, provide partial local solutions but do not fully capture the global competitive nature of multi-agent racing.

Towards this end, we propose \textbf{$\alpha$-RACER} ($\alpha$-potential function based \textbf{R}eal-time \textbf{A}lgorithm for \textbf{C}ompetitiv\textbf{E} \textbf{R}acing). We study the following key question:
\begin{quote}
    How to design real-time algorithms for  autonomous racing that approximate Nash equilibrium, while accounting for nonlinear tire forces, long-horizon planning and accommodating competitive interaction in any number of competing vehicles?
\end{quote}
We highlight three main contributions of this work towards answering the question posed above.

\noindent\textbf{1. \underline{Modeling Contribution}:} We model the multi-agent interaction as an infinite-horizon discounted dynamic game and introduce a novel policy parametrization to enable competitive maneuvers. Specifically, we propose MPC-based policies that track a specially designed, parameterized \textit{reference trajectory}, while avoiding other vehicles. This trajectory is derived by adjusting the optimal single-agent race-line to account for the presence of other agents, enabling competitive racing maneuvers such as overtaking and blocking. Additionally, we structure each agent’s immediate utility function to increase its relative progress along the track in comparison to other at each time-step.
Moreover, our approach is modular with respect to the vehicle dynamics model. In this work, we use a nonlinear vehicle dynamics model that captures nonlinear tire forces, but the framework can also incorporate other advanced racing technologies, such as ``push-to-pass''. By designing the reference trajectory in this way, we integrate long-term strategic planning for optimizing lap times with tactical planning for effective competition with other cars.

\noindent\textbf{2. \underline{Algorithmic Contribution}:}
We present an algorithmic approach to compute Nash equilibrium of the dynamic game by leveraging the recent framework of dynamic \(\alpha\)-potential functions (developed in \citep{guo2023markov, maheshwari2024decentralized, guo2024alpha}). Specifically, we compute an \(\alpha\)-potential function that captures the change in each agent's (long-term) value function resulting from unilateral deviations in its policy parameters. The structure of \(\alpha\)-potential function allows us to approximate the Nash equilibrium as an optimizer of this function (refer Proposition \ref{prop: NE_def_PotMax}).
We leverage this structure to develop a two-phase algorithmic approach to approximate the Nash equilibrium, consisting of an \textit{offline} and an \textit{online} phase. In the offline phase, we learn an \(\alpha\)-potential function using simulated game data. During the online phase, the agent observes the current state of the game and selects the policy parameters that maximize the potential function in that state. This approach enables autonomous agents to engage in competitive maneuvers with reduced computational demands.

\noindent \textbf{3. \underline{Numerical Evaluation}:} We numerically validate the effectiveness of our approach by applying it to three-car autonomous racing scenarios. Our results show that the approximation gap of our learned potential function is small, enabling us to closely approximate Nash equilibrium strategies. We demonstrate that the maximizer of the potential function effectively captures competitive racing maneuvers. Furthermore, we show that our method outperforms opponent policies—obtained using iterated best response, and finite-horizon self-play reinforcement learning—in most cases.

\paragraph{Other Related Works.} Several works in the autonomous racing literature use the solution concept of (open-loop) \textit{generalized Nash equilibrium} (e.g., \citep{wang2019bgame, jia2023rapid}) to incorporate hard collision avoidance constraints, which introduces added computational complexity. {In contrast, our modeling framework incorporates collision avoidance in two ways: \((i)\) through the MPC controller used to track the reference trajectory, and \((ii)\) by slowing down vehicles when they enter a pre-defined radius around each other.}

Our algorithmic approach also contributes to the growing literature on computational game theory for multi-robot systems \citep{laine2021multi, cleac2019algames, fridovich2020efficient, liu2023learning, peters2024contingency, kavuncu2021potential, sadigh2016planning, wang2020game} by offering a new method to compute approximate Nash equilibrium.

\paragraph{Organization.} In Section \ref{sec: Model}, we describe our model of multi-car racing as a dynamic game, including the policy parametrization, utility function, and vehicle dynamics model. In Section \ref{sec: Approximation}, we present our algorithmic approach. In Section \ref{sec: Numerical}, we evaluate the performance of our approach using a numerical racing simulator. Finally, we conclude with remarks in Section \ref{sec: Conclusion}.

% \begin{table}[h!]
%     \centering
%     \begin{tabular}{|c|c|c|c|c|}
%         \hline
%         \textbf{Ref.} & \textbf{Vehicle Model} & \textbf{Game Model} &  \textbf{\# of cars} & \textbf{Planning}  \\ \hline
% \cite{buyval2017deriving} & Dynamic & Known Dynamic Obstacle & \(2\) & Local
%         \\
% \hline\cite{jung2021game} & Dynamic & Stackelberg Game & \(\geq 2\)& Local\\ 
% \hline 
% \cite{notomista2020enhancing} & Kinematic & Zero-sum (IBR)  & 2& Local
% \\
% \hline 
% \cite{wang2019bgame} & Kinematic & Zero-sum (IBR) Game & \(2\)& Local
% \\
% \hline 
% \cite{wang2021game} & Kinematic & Non-zero sum game & \(\geq 2\)& Local
% \\
% \hline 
% \cite{liniger2019noncooperative} & Kinematic & Stackelberg/Nash Game & 2& Local
% \\
% \hline 
% \cite{jia2023rapid}  & Kinematic & Potential Game & \(\geq 2\)& Local
% \\ \hline
% \cite{williams2017autonomous} & Kinematic & IBR & \(\geq 2\) & Local
% \\
% \hline
% \cite{sinha2020formulazero} & Kinematic & Robust RL / Self play & \(\geq 2\) & Local 
% \\
% \hline
% \cite{schwarting2021stochastic} & Kinematic & iLQG & \(\geq 2\) & Local 
% \\
% \hline
% \cite{thakkar2022hierarchical}  &  Dynamic & Zero-sum & 2& Global
% \\
% \hline 
% \cite{kalaria2023towards}&  Dynamic & Zero-sum & 2& Global
% \\
% \hline
% \cite{zhu2023sequential, zhu2024sequential} & Dynamic & General-sum & \(\geq 2\) & Local
% \\
% \hline
% \textbf{Ours} & Dynamic & General-sum & \(\geq 2\) & Global
% \\
% \hline
%     \end{tabular}
%     \caption{Comparison to previous literature on game theoretic planning for multi-car autonomous racing.}
%     \label{tab:sample_table}
% \end{table}

\section{Modeling Multi-car Autonomous Racing}\label{sec: Model}
In this section, we first present the necessary preliminaries on dynamic game theory, followed by a novel model of multi-car autonomous racing as an instantiation of a dynamic game.
\subsection{Preliminaries on Dynamic Game Theory}\label{ssec: Preliminaries}
Consider a game involving \(N\) strategic players, with the set of players denoted as \(\mc{N} := \{1, 2, \dots, N\}\). The game proceeds in discrete time steps, indexed by \(t \in \mathbb{N}\). 
At each time step \(t\), the state of player \(i \in \mc{N}\) is represented as \(\mathbf{x}_t^i \in \mc{X}^i\), where \(\mc{X}^i\) is the set of all possible states for player \(i\). The joint state of all players at time \(t\) is denoted by \(\mathbf{x}_t = \left( \mathbf{x}_t^i\right)_{i\in \mc{N}} \in \mc{X}\), where \(\mc{X} := \times_{i\in \mc{N}} \mc{X}^i\).
At every time step \(t\), each player \(i \in \mc{N}\) selects an action \(\mathbf{u}_t^i \in \mc{U}^i\), where \(\mc{U}^i\) is the set of feasible actions for player \(i\). The joint action at time \(t\) is expressed as \(\mathbf{u}_t = \left( \mathbf{u}_t^i\right)_{i\in \mc{N}}  \in \mc{U}\), where \(\mc{U} := \times_{i\in \mc{N}} \mc{U}^i\).

For each player \(i \in \mc{N}\), given the current state \(\mathbf{x}_t^i\) and the joint action \(\mathbf{u}_t\), the state transitions to a new state at time \(t+1\) according to the dynamics \(\mathbf{x}_{t+1}^i = f^i(\mathbf{x}_t, \mathbf{u}_t^i),\)
where \(f^i: \mc{X} \times \mc{U}^i \to \mc{X}^i\) describes the state transition dynamics of player \(i\). 
Finally, at each time step \(t\), player \(i \in \mc{N}\) receives a reward \(r^i_t(\mathbf{x}_t, \mathbf{u}_t, \mathbf{x}_{t+1})\), which depends on the current and next joint state, and joint action of all players.
We assume that players select their actions based on a state feedback strategy, denoted by \(\pi^i: \mc{X} \to \mc{U}^i\), such that \(\mathbf{u}_t^i = \pi^i(\mathbf{x}_t)\).  The set of strategies of player \(i \in \mc{N}\) is denoted by \(\Pi^i\), and the set of joint strategies by \(\Pi = \times_{i\in\mc{N}} \Pi^i\).
In our study, we consider a parameterized set of policies. Specifically, for any player \(i \in \mc{N}\), the player adopts a policy \(\pi^i(\cdot;\theta^i) \in \Pi^i\), where \(\theta^i \in \Theta^i\) represents the policy parameter.
We consider non-myopic players who aim to maximize their long-term discounted utility starting from any state \(\mathbf{x} \in \mc{X}\). Specifically, each player \(i \in \mc{N}\) selects a parameter \(\theta^i \in \Theta^i\) to optimize the discounted infinite-horizon objective, 
\(
    V^i(\mathbf{x}, \theta^i, \theta^{-i}) := \mathbb{E}\left[ \sum_{t=0}^{\infty} \gamma^t r^i_t(\mathbf{x}_t, \mathbf{u}_t, \mathbf{x}_{t+1}) \right],
\)
where \(\gamma \in [0,1)\) is the discount factor, \(\mathbf{x}_0 = \mathbf{x}\), \(\mathbf{u}_t^i = \pi^i(\mathbf{x}_t; \theta^i)\), \(\mathbf{u}_t^{-i} = \pi^{-i}(\mathbf{x}_t; \theta^{-i})\)\footnote{We use the standard game theory notation of \(\pi^{-i}\) (resp. \(\theta^{-i}\)) to denote the joint policy (joint policy parameters) of all players except player \(i\).}, and the state transitions according to \(\mathbf{x}_{t+1}^i = f^i(\mathbf{x}_t^i, \mathbf{u}_t^i)\).

\begin{definition}\label{def: NE}
    For every \(\mathbf{x} \in \mc{X}\), a joint strategy profile \(\theta^\ast \in \Theta\) is called a \emph{\(\epsilon\)-Nash equilibrium} if, for every \(i \in \mc{N}, \theta^i\in \Theta^i\),
\(
    V^i(\mathbf{x}, \theta^{\ast,i}, \theta^{\ast, -i}) \geq V^i(\mathbf{x}, {\theta}^i, \theta^{\ast, -i}) - \epsilon. \)
If \(\epsilon = 0\), the strategy profile \(\theta^\ast\) is referred to simply as a Nash equilibrium. 
\end{definition}
\subsection{Multi-car Racing as a Dynamic Game}\label{ssec: RacingAsGame}
In this subsection, we formulate the multi-car racing problem as a dynamic game by detailing the various components of the game.

\vspace{0.1cm}
\noindent \textbf{\underline{Set of Players}.} Each car is modeled as a strategic player in the dynamic game.

\vspace{0.1cm}
\noindent \textbf{\underline{Set of States and Actions}.} For every car \(i \in \mc{N}\), the state \(\mathbf{x}^i=(p_x^i, p_y^i, \phi^i, v_x^i, v_y^i, \omega^i)\), where \((p_x^i, p_y^i)\) denote the longitudinal and lateral position of car \(i\) in the \textit{Frenet frame} along the track; \(\phi^i\) denotes the orientation of the car in the {Frenet frame} along the track; {\((v_x^i, v_y^i)\) denote the longitudinal and lateral velocities of car \(i\) in the {Frenet frame}}; and \(\omega^i\) denotes the angular velocity of car \(i\) in the body frame. Additionally, \(\mathbf{u}^i = (d^i, \delta^i)\), where \(\delta^i \in [\delta_{\min}, \delta_{\max}]\) denotes the steering angle of car \(i\) and \(d^i \in [d_{\min}, d_{\max}]\) is the throttle input of car \(i\). 

\vspace{1mm}
\noindent \textbf{\underline{Dynamics}.} The most widely used dynamics models in the context of racing are the \textit{kinematic} \citep{jia2023rapid, notomista2020enhancing, wang2019game, wang2021game} and \textit{dynamic bicycle models} \citep{Kalaria2023AdaptivePA, Brunke2020LearningMP}. In this work, we use the dynamic model as it can accurately model the high-speed maneuvers of the car. A detailed vehicle model description is in Supplementary. 
On top of the standard dynamic bicycle model, we also incorporate near-collision behavior in our dynamics. Suppose two cars, $i$ and $j$, are within an unsafe distance from each other and $p_{x,t}^{i} > p_{x,t}^{j}$, we reduce their velocities to replicate the time lost due to collision in an actual race with more penalty for the car behind than the one ahead. More concretely, we update the dynamics as $v_{x,t+1}^{i} = (1/2)\cdot v_{x,t}^{i}$ and $v_{x,t+1}^{j} = (1/3)\cdot v_{x,t}^{j}$. Moreover, if a car goes out of track boundary, i.e., when $|p_{y,t}^i|>{w_{\max}}/{2}$ (where \(w_{\max}\) is the width of track) we penalize it by reducing its speed and re-align the car along the track. More concretely, we update the dynamics as $v_{x,t+1}^i = v_{x,t}^i/2$ and $\phi_{t+1}^i=0$.

\vspace{1mm}
\noindent 
\textbf{\underline{Policy Parametrization}.} In this section, we introduce a \emph{novel} policy parametrization \((\Theta^i)_{i\in\mc{N}}\) designed to capture competitive driving behaviors, such as overtaking, blocking, apex hugging, late braking, and early acceleration. To achieve this, we restrict the set of policies to MPC controllers that track a \textit{reference trajectory} specifically designed to encode such competitive behaviors in multi-car racing. For each car \(i \in \mc{N}\), the parametrization of the MPC controller and its reference trajectory is represented by \(\theta^i\), which is characterized by five variables: \((q^i, \zeta^i, s_1^i, s_2^i, s_3^i)\). Before describing these parameters in detail, we present the MPC controller.

At each time step \(t\), for every car \(i \in \mc{N}\), the MPC controller is determined by solving an optimization problem (cf. \eqref{eqn:opt}) over a planning horizon of \(K\) steps, indexed by \(k\). The optimization is parameterized by: \((a)\) the longitudinal and lateral positions on the reference trajectory (to be defined shortly), denoted by \((p_{x|t}^{\textsf{ref},i,k},p_{y|t}^{\textsf{ref},i,k})_{k\in [K]}\); \((b)\) the current state of car \(i\) at time \(t\), \(\mathbf{x}_t^i\); and \((c)\) the longitudinal and lateral positions of the opponent cars located just behind and just ahead of car \(i\). We use the notation \(j^\ast, j_\ast \in \mc{N}\) to denote the car in front and the car behind car \(i\) at the start of the planning window, based on longitudinal position. For any opponent vehicle \( j \in \{ j^\ast, j_\ast \} \), we assume that the lateral velocity is zero during the planning horizon for the MPC controller, and the longitudinal velocity remains constant at its value at the start of the planning horizon. That is, for every \(k\in [K]\), \(p_{y|t}^{j,k} = p_{y,t}^{j}\), and \(p_{x|t}^{j,k+1} = p_{x|t}^{j,k} + \Delta t \cdot v_{x,t}^{j}\).
% ,where \(p_{x,t}^{j}\), \(p_{y,t}^{j}\), and \(v_{x,t}^{j}\) represent the longitudinal position, lateral position, and longitudinal velocity of the opponent car.
% , respectively, in the Frenet frame. 
With this setup, we can now describe the MPC optimization problem:
\vspace{-0.2cm}
\begin{subequations} \label{eqn:opt}  
\begin{align}
\min_{(\textbf{x}_t^{i,k})_{k=1}^{K},(\textbf{u}_t^{i,k})_{k=0}^{K-1}} &\sum_{k=1}^K \left\| \begin{pmatrix} p_{x|t}^{i,k} - p_{x|t}^{\textsf{ref},i,k} \\ p_{y|t}^{i,k} - p_{y|t}^{\textsf{ref},i,k} \end{pmatrix} \right\|_{\textbf{Q}}^2 
+ \sum_{k=1}^{K-1} \left\| \begin{pmatrix} d_t^{i,k} - d_t^{i,k-1} \\ \delta_t^{i,k} - \delta_t^{i,k-1} \end{pmatrix} \right\|_{\textbf{R}}^2  \label{eq: MPCObjective}\\
\text{s.t.  } &\mathbf{x}_{t}^{i,k+1} = f^i (\mathbf{x}_t^{k},\mathbf{u}_t^{i,k}), \quad \forall k = 0,1,..., K-1,  \label{eq: MPCDynamics}\\
\text{           } &\mathbf{x}_t^{i,0} = \mathbf{x}_t^i, \label{eq: MPCInitialCondition}\\
\text{           } &\Delta \delta_{\min}^i \le \delta_t^{i,k} - \delta_t^{i,k-1} \le \Delta \delta_{\max}^i, \quad \forall k = 0,1,..., K-1, \label{eq: MPCControlConstraints}\\
&d_{\min}^i \le d_t^{i,k} \le d_{\max}^i, \quad \forall k = 0,1,..., K-1, \label{eq: MPCThrottleConstraints}\\
& |p_{y,t}^{i,k}| \leq w_{\max}/2, \quad \forall k = 1,..., K, \label{eq: MPCStateConstraintTrack}\\
& |p_{y|t}^{i,k} - p_{y|t}^{j,k}| \geq p_y^{\min}, \quad \forall k = 1,..., K, \forall j \neq i, \label{eq: MPCLateralCollisionAvoidance}\\
& |p_{x|t}^{i,k} - p_{x|t}^{j,k}| \geq p_x^{\min}, \quad \forall k = 1,..., K, \forall j \neq i, \label{eq: MPCLongitudinalCollisionAvoidance} 
\end{align}
\end{subequations}
where \(\mathbf{x}_t^{i,k} = (p_{x|t}^{i,k}, p_{y|t}^{i,k}, \phi_{t}^{i,k}, v_{x|t}^{i,k}, v_{y|t}^{i,k}, \omega_t^{i,k})\) is the state of car \(i\) in the Frenet frame at the \(k^{\textsf{th}}\) step in the planning horizon; \(\mathbf{u}_t^{i,k} = (d_t^{i,k}, \delta_{t}^{i,k})\) is the control input of car \(i\) at the \(k^{\textsf{th}}\) step in the planning horizon; \(w_{\max}\) is the track length, and \(p_x^{\min}\) and \(p_y^{\min}\) are the minimum required separation between two cars in the longitudinal and lateral directions, respectively; \(d_{\min}^i\) and \(d_{\max}^i\) are the throttle limits, and \(\Delta \delta_{\min}^i\) and \(\Delta \delta_{\max}^i\) are the steering rate limits; and $\textbf{Q}$ and $\textbf{R}$ are positive definite matrices. Following the MPC approach, the control input at time \(t\) is then \(\mathbf{u}_t^{i,0}\).

In \eqref{eqn:opt}, \eqref{eq: MPCObjective} defines the MPC objective, where the first term penalizes the tracking error relative to the reference trajectory, and the second term penalizes variations in the control input. \eqref{eq: MPCDynamics} represents the system dynamics constraint, while \eqref{eq: MPCInitialCondition} ensures the planning horizon begins at the car's current state. \eqref{eq: MPCControlConstraints} and \eqref{eq: MPCThrottleConstraints} enforce constraints on the control inputs, including throttle limits and steering rate bounds. \eqref{eq: MPCStateConstraintTrack} ensures the car stays within the track boundaries, and \eqref{eq: MPCLateralCollisionAvoidance} and \eqref{eq: MPCLongitudinalCollisionAvoidance} enforce the minimum separation between vehicles in the lateral and longitudinal directions, respectively. Next, we describe the policy parameter \(\theta^i = (q^i, \zeta^i, s_1^i, s_2^i, s_3^i) \in \mathbb{R}^5\), which parametrizes the policy.

\((i)\) \textbf{Parameter \(q^i\):} In \eqref{eq: MPCObjective}, we take $\textbf{R}=\textbf{I}$ and $\textbf{Q}=q^i\cdot \textbf{I}$, where \(\mathbf{I}\) is the identity matrix. A higher \(q^i\) value results in a more aggressive controller that closely follows the racing line, but it can introduce oscillations that may increase lap time. In contrast, a lower \(q^i\) value allows for smoother merging with reduced oscillations, though it may result in larger lateral errors, increased time loss at corners, and a higher risk of track boundary violations.

\textit{(ii)} \textbf{Parameter \(\zeta^i\):} We use this parameter to develop a perturbed version of the optimal (single-agent) race-line (see Supplementary for a discussion on race-line), which is generated by sampling points along the optimal racing line at time intervals of \(\Delta t\). More formally, let the optimal race-line be denoted by \(\mathbf{x}^{\textsf{rl},i}\). Given the current state \(\mathbf{x}_t^i\), we find the closest point on the race-line and consider a race-line starting at time \(t\), denoted by \(\bar{\mathbf{x}}_t^{\textsf{rl},i}\). Using this, we compute a trajectory of length \(K\), denoted by \((p_{x|t}^{\textsf{pert},i,k}, p_{y|t}^{\textsf{pert},i,k})_{k \in [K]}\). Specifically, for every \(k \in [K]\), we construct:
\[
p_{x|t}^{\textsf{pert},i,k} = p_{x|t}^{\textsf{pert},i,k-1} + v_{x|t}^{\textsf{pert},i,k-1} \cdot \Delta t, \quad p_{y|t}^{\textsf{pert},i,k} = p_{y|t}^{\textsf{pert},i,k-1} + v_{y|t}^{\textsf{pert},i,k-1} \cdot \Delta t,
\]
where \(v_{x|t}^{\textsf{pert},i,k-1} = \zeta^i \bar{v}_{x,t+k-1}^{\textsf{rl},i}\) and \(v_{y,t}^{\textsf{pert},i,k-1} = \zeta^i \bar{v}_{y,t+k-1}^{\textsf{rl},i}\) are the perturbed race-line velocities. Higher values of \(\zeta^i\) capture how aggressively the vehicle wants to follow the optimal race-line.

\textit{(iii)} \textbf{Parameters \(s_1, s_2, s_3\):}  
% We define \((p_{x,t}^{\textsf{ref},i,k}, p_{y,t}^{\textsf{ref},i,k})\) is the reference trajectory for car \(i\), which is specially designed to capture the competitive racing behavior. 
    The reference trajectory, \((p_{x|t}^{\textsf{ref},i,k}, p_{y|t}^{\textsf{ref},i,k})\), is generated by modifying the perturbed race-line  \((p_{x|t}^{\textsf{pert},i,k}, p_{y|t}^{\textsf{pert},i,k})_{k\in[K]}\) by accounting for the positions and velocities of the cars immediately ahead (in terms of longitudinal coordinates) of the ego car and immediately behind (in terms of longitudinal coordinates) ego car. Let's denote the ego car by \(i\), the car immediately ahead of this car by \(j^\ast\), and the one immediately behind by \(j_\ast\). We define the reference trajectory as follows \(p_{y|t}^{\textsf{ref}, i,k} = \textsf{clip}({p}_{y|t}^{\textsf{pert}, i, k} + {p}_{y|t}^{\textsf{ot}, i, k} +  {p}_{y|t}^{\textsf{bl},i,k}, [-w_{\max},w_{\max}]),\)
    where 
\begin{align*}
{p}_{y|t}^{\textsf{ot},i,k} & = \textsf{sign}({p}_{y,t}^i - {p}_{y,t}^{j^\ast})
            \max\left\{(s_1-  |({p}_{y,t}^i - {p}_{y,t}^{j^\ast})|) \exp\left(-s_2 \left(\Delta p_{x|t}^{i,j^\ast,k}\right)^2 \right)  , 0\right \}  \\ 
            & \quad + \textsf{sign}({p}_{y,t}^i - {p}_{y,t}^{j_\ast})
            \max\left\{(s_1- |({p}_{y,t}^i - {p}_{y,t}^{j_\ast})|)\exp\left(-s_2 \left(\Delta p_{x|t}^{i,j_\ast,k}\right)^2 \right) , 0\right \}, 
    \end{align*}
    is an adjustment for overtaking that smoothly changes the trajectory of the ego vehicle opposite the leading vehicle to overtake. Here, for any \(j\in \{j^\ast, j_\ast\}\),
 \(\Delta p_{x|t}^{i,j,k} 
 = p_{x|t}^{\textsf{pert},i,k} - p_{x|t}^{j,k}\), and
\begin{align*}
    \bar{p}_{y|t}^{\textsf{bl},i,k} &= \mathbbm{1}(v_{x|t}^{\textsf{pert},i,k}\leq v_{x,t}^{j_\ast})\mathbbm{1}(p_{x|t}^{\textsf{pert},i,k}\geq p_{x|t}^{j_\ast,k}) h(p_{y|t}^{j_\ast,k}, p_{x|t}^{\textsf{pert},i,k},v_{x|t}^{\textsf{pert},i,k},v_{x,t}^{j_\ast},\Delta p_{x|t}^{i,j_\ast,k})\\
    &+\mathbbm{1}(v_{x|t}^{\textsf{pert},i,k}\leq v_{x,t}^{j^\ast})\mathbbm{1}(p_{x|t}^{\textsf{pert},i,k}\geq p_{x|t}^{j^\ast,k}) h(p_{y|t}^{j^\ast,k}, p_{x|t}^{\textsf{pert},i,k},v_{x|t}^{\textsf{pert},i,k},v_{x,t}^{j^\ast},\Delta p_{x|t}^{i,j^\ast,k})
\end{align*}
\noindent is the adjustment for blocking that smoothly changes the trajectory of the ego vehicle towards trailing vehicle to block it. Here, for any \(j\in \{j^\ast, j_\ast\}\),
\begin{align*}
&h(p_{y|t}^{j,k}, p_{x|t}^{\textsf{pert},i,k},v_{x|t}^{\textsf{pert},i,k},v_{x,t}^{j},\Delta p_{x|t}^{i,j,k}) \\ &\quad = (p_{y|t}^{j,k} - p_{y|t}^{\textsf{pert},i,k})(1-\exp(-s_3(v_{x|t}^{\textsf{pert},i,k}- v_{x,t}^{j})))\exp(-s_2(\Delta p_{x|t}^{i,j,k})^2).
\end{align*} 
 %    \begin{align*}
 %    \bar{p}_{y,k}^{i,\textsf{block}} = \begin{cases}
 %             -\max\left\{s_3 \exp\left(-s_4 \left(\frac{\bar{p}_{x,0}^{j_\ast}-\bar{p}_{x,0}^{i,\textsf{rl}}}{\bar{v}_{x,0}^{j_\ast}-\bar{v}_{x,0}^{i,\textsf{rl}}}- k \Delta t\right)^2 \right) - (\bar{p}_{y,0}^i - \bar{p}_{y,0}^{j_\ast}) , 0\right \} & \text{if} \ \bar{p}_{y,0}^i \geq \bar{p}_{y,0}^{j_\ast} \\ 
 %            \max\left \{s_3 \exp\left(-s_4 \left(\frac{\bar{p}_{x,0}^{j_\ast}-\bar{p}_{x,0}^{i,\textsf{rl}}}{\bar{v}_{x,0}^{j_\ast}-\bar{v}_{x,0}^{i,\textsf{rl}}}- k \Delta t\right)^2 \right) - (\bar{p}_{y,0}^{j_\ast} - \bar{p}_{y,0}^i) , 0\right \} & \text{if} \ \bar{p}_{y,0}^i \leq \bar{p}_{y,0}^{j_\ast},
 %        \end{cases} 
 %    \end{align*}
 % where \(j_\ast\) is the nearest car behind car \(i\).  
    % \item[(iii)]  
    % \item[(iv)] \(p_x^{\min}\) and \(p_y^{\min}\) are the minimum required separation between two cars in the longitudinal and lateral directions, respectively. 

\noindent 
\textbf{\underline{One-step Utility Function}.} 
We consider that the one-step utility for every car is to maximize progress along track: \(
\mathbb{R}\ni r^i(\mathbf{x}_t,\mathbf{u}_t,\mathbf{x}_{t+1}) = (p_{x,t+1}^{i} - \max_{j\neq i} p_{x,t+1}^{j})- (p_{x,t}^{i} - \max_{j\neq i} p_{x,t}^{j}). \)

\section{Approximating Multi-agent Interaction}\label{sec: Approximation}
In this section, we provide a tractable approach to compute an approximate Nash equilibrium for the racing game described in Section \ref{sec: Model}. Core to our approach is the framework of \(\alpha\)-potential functions, recently introduced in \citep{guo2023markov,maheshwari2024decentralized, guo2024alpha}. 

\subsection{\(\alpha\)-Potential Function}

\begin{definition}\label{def: NearPotentialFunction}
    A potential function \(\Phi: \mc{X}\times \Theta \rightarrow \mathbb{R}\) is called a \emph{dynamic \(\alpha\)-potential function}\footnote{For convenience, we adopt a slightly different definition of the \(\alpha\)-potential function than that in \cite{guo2023markov, maheshwari2024decentralized}; however, the results of this work extend to the definitions used there.} with \emph{approximation parameter} \(\alpha\) if for every \(\mathbf{x}\in \mc{X}, i\in \mc{N}, \theta\in \Theta, \theta^{i}{}'\in \Theta^i\),  
    \begin{align}\label{eq: NearMPGDef}
        |(\Phi(\mathbf{x}, \theta^i, \theta^{-i})-\Phi(\mathbf{x}, \theta^i{}', \theta^{-i}) ) - (V^i(\mathbf{x}, \theta^i, \theta^{-i})-V^i(\mathbf{x}, \theta^i{}', \theta^{-i}) )| \leq \kappa.  
    \end{align}
\end{definition}
This definition intuitively requires that for any agent, the change in its value function resulting from a unilateral adjustment to its policy parameter can be closely approximated by the corresponding change in the dynamic \(\alpha\)-potential function. This property allows us to approximate the Nash equilibrium as an optimizer of the near-potential function (as proved in the supplementary material). 

\vspace{-1mm}
\begin{proposition}\label{prop: NE_def_PotMax}
    Given an $\alpha$-potential function $\Phi$, for any $\mathbf{x} \in \mathcal{X}$, $\lambda > 0$, and any policy $\theta^\ast$ satisfying
\(
\Phi(\mathbf{x}, \theta^\ast) \geq \max_{\theta \in \Theta} \Phi(\mathbf{x}, \theta) - \lambda,\)
the policy $\theta^\ast$ constitutes a $(\lambda + \alpha)$-approximate Nash equilibrium.
\end{proposition}

\vspace{-7mm}
\subsection{Computational Approach}
\vspace{-1mm}
Our approach for real-time approximation of Nash equilibrium relies on two phases: offline and online. In the offline phase, we learn an \(\alpha\)-potential function using simulated game data. In the online phase, the ego vehicle updates its policy parameters by optimizing the potential function. 

 \vspace{1mm}
 \noindent \underline{\textbf{Offline Phase:}} 
    We parameterize the potential function through using a feed-forward neural network with ReLU activation and a BatchNorm layer added at the beginning. More concretely, we define the parametrized potential function as \(\Phi(\cdot;\phi): \mc{X} \times A \rightarrow \mathbb{R}\), where \(\phi\) denotes the weights of neural network. Using Definition \ref{def: NearPotentialFunction}, we cast the problem of learning potential function as a semi-infinite program as shown below: 
\begin{align} 
    \min_{\substack{ y, \phi}}\quad  &  y\label{eq: LinearSemiInfiniteFormulation}\\ 
    \text{s.t.}\quad &  \Big| (\Phi(\mathbf{x}, \theta^i,\theta^{-i}; \phi) -\Phi(\mathbf{x},\theta^i{}',\theta^{-i}; \phi)) - (V^i(\mathbf{x},\theta^i,\theta^{-i};\upsilon) -V^i(\mathbf{x},\theta^i{}',\theta^{-i};\upsilon)) \Big| \leq y, \notag \\  &\quad ~\forall i \in I, ~ \forall \theta^i, \theta^i{}' \in \Theta^i, ~ \forall \theta^{-i}\in \Theta^{-i}, \mathbf{x}\in\mc{X}, \notag 
\end{align}
where we use a neural network (same architecture as potential function), with parameters \(\upsilon\), to estimate the value function \(V^i\) for every \(i\in \mc{N}\). Let \(\phi^\ast\) be a solution of the above optimization problem. 
The main challenge with solving \eqref{eq: LinearSemiInfiniteFormulation} is that the there are un-countably many constraints, one constraint corresponding to each value of initial state and policy parameter. Therefore, we numerically solve \eqref{eq: LinearSemiInfiniteFormulation} by using  simulated game data with randomly chosen starting position and policy parameters. Details of simulated game are discussed in next section.  

\vspace{1mm}
\noindent \underline{\textbf{Online Phase:}} Leveraging Proposition \ref{prop: NE_def_PotMax}, the ego vehicle optimizes the learned potential function, i.e. \(\Phi(\cdot, \phi^\ast)\), to approximate the Nash equilibrium policy parameter. More formally, given the current state \(\mathbf{x}_t\), the ego vehicle computes 
\(
    \theta^\ast \in \arg\max_{\theta\in\Theta}\Phi(\mathbf{x}_t, \theta;\phi^\ast),
\)
using a non-linear optimization solver. 
Using \(\theta^\ast\), the ego vehicle takes action \(\mathbf{u}_t^i = \pi^i(\mathbf{x}_t; \theta^{\ast, i})\).

% \begin{align*}
%     \textsf{Regret} =  \max_{i\in [N]}|\max_{ \theta_i} V_i(x,\theta_{i}',\theta^\ast_{-i}) - V_i(x,\theta^\ast_i,\theta^\ast_{-i}) |
% \end{align*}

\vspace{-4mm}
\section{Numerical Evaluation}\label{sec: Numerical}
\vspace{-1mm}
Here, we evaluate our approach on  a numerical simulator by focusing on three questions: \textbf{(Q1)} Can our approach closely approximate Nash equilibrium and generate competitive behavior? \textbf{(Q2)} How do hyper-parameters like the discount factor $\gamma$ and the amount of data used to learn the \(\alpha\)-potential function affect the performance? \textbf{(Q3)} How does our approach compare against common baselines?

\vspace{1mm}
\noindent \textbf{Experimental Setup:} We generate a dataset of 4000 races, each lasting 50 seconds, conducted with randomly chosen policy parameters and involving 3 cars. This dataset is used to first train value function estimators $V^1$, $V^2$, and $V^3$ for each of the cars. These are then used to learn $\Phi$ using \eqref{eq: LinearSemiInfiniteFormulation}. To maximize the learned potential function, we use gradient ascent with a learning rate of $10^{-4}$ and warm-start by using the solution from the previous time step.

% FIGS HERE
\begin{figure}
    \centering
    \begin{subfigure}{} 
        \includegraphics[width=5cm]{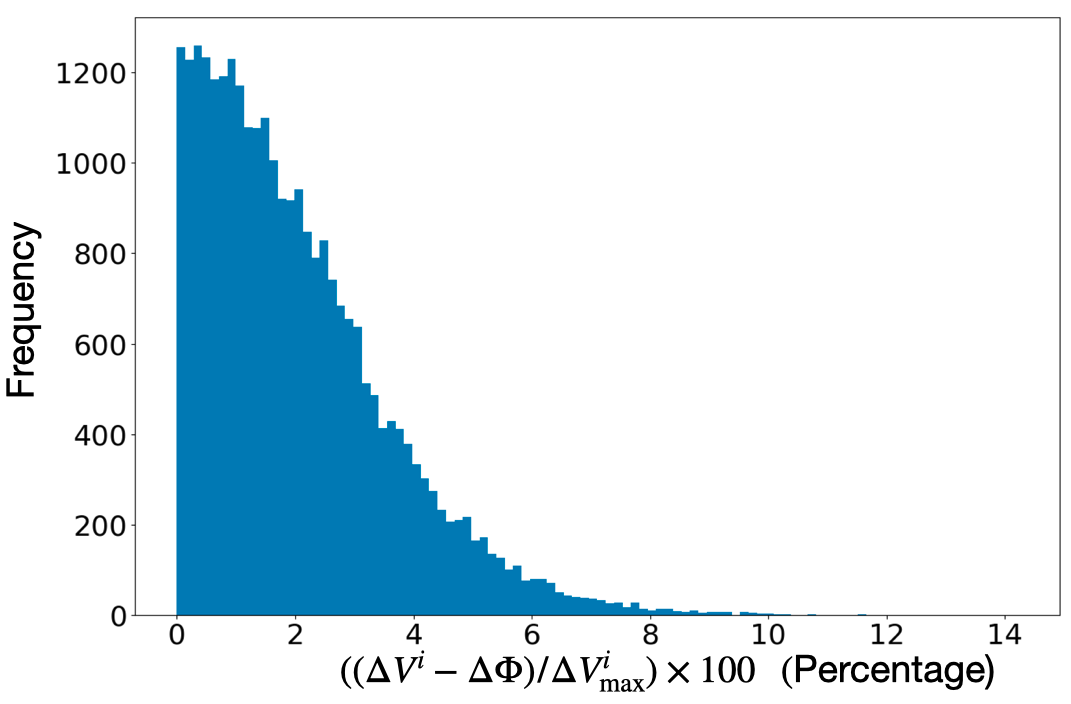}
        % {Figures/ds_new.png}
        \label{subfig:ds}
    \end{subfigure}
    \begin{subfigure}{} 
        \includegraphics[width=4.3cm]{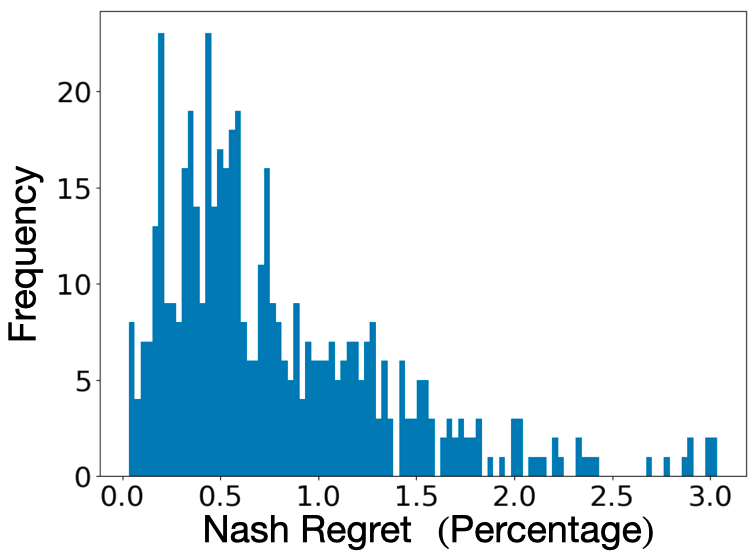}
        % {Figures/regrets_new.png}  
        \label{subfig:regrets}
    \end{subfigure}
    \caption{Histogram of (a) Relative approximation gap of potential function (b) Nash regret}
    \label{fig:regrets}
\vspace{-7mm}  
\end{figure}

\vspace{-5mm}
\subsection{Competitive Behavior by Approximating Nash Equilibrium}
\vspace{-2mm}
We observe that the approximation gap of the learned potential function is small. As shown in Figure~\ref{fig:regrets}(a), the approximation gap across all states and policy parameters used in the training samples remains within 10\% of the value function's range, with a median gap of approximately 2\%.  
Next, we demonstrate that the optimization solver effectively computes the maximizer of the potential function, leading to a lower Nash approximation error. In particular, Figure~\ref{fig:regrets}(b) shows the \textit{Nash regret} for the ego agent, defined as  
\(
\max_{\theta^i} V^i(\mathbf{x}, \theta^i, \theta^{\ast,-i}) - V^i(\mathbf{x}, \theta^{\ast,i}, \theta^{\ast,-i}),
\)  
where \(\theta^\ast\) is the optimizer of the potential function with the starting state \(\mathbf{x}\). The regret is plotted for different game states during a race, and we observe that it remains within 3\% of the range of the value function.  
In summary, Figure~\ref{fig:regrets} highlights that the dynamic game admits an \(\alpha\)-potential function with small \(\alpha\), and that we accurately compute a near-optimal solution to the potential function.

\begin{figure}
    \centering
    \begin{subfigure}{} 
        \includegraphics[width=5.5cm]{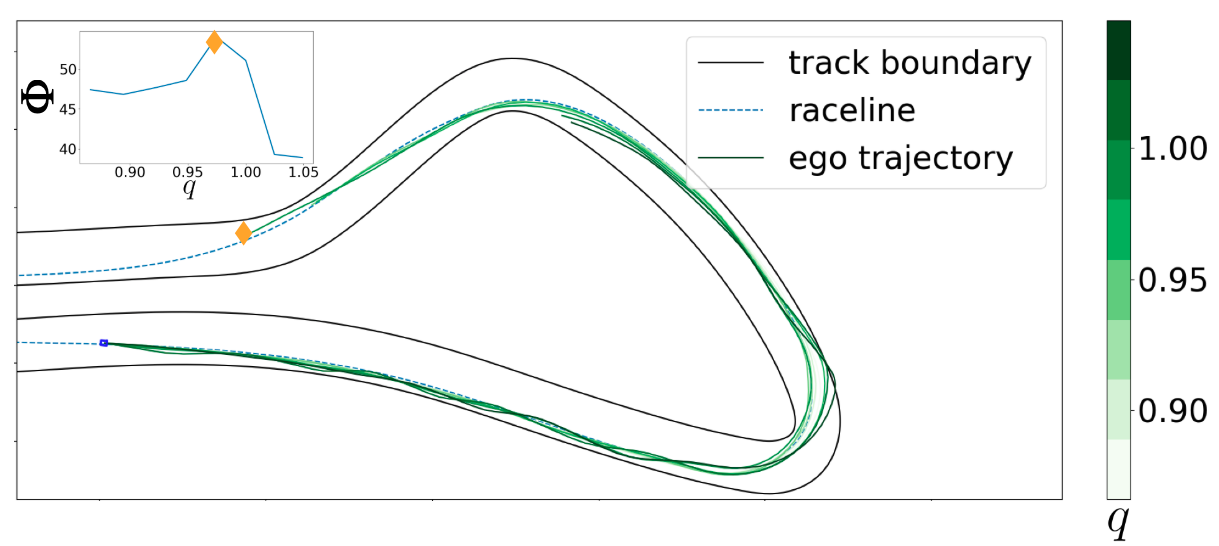}
        % {Figures/p_q_new.png}    
        % \label{subfig:p_q}
    \end{subfigure}
    \begin{subfigure}{} 
        \includegraphics[width=5.5cm]{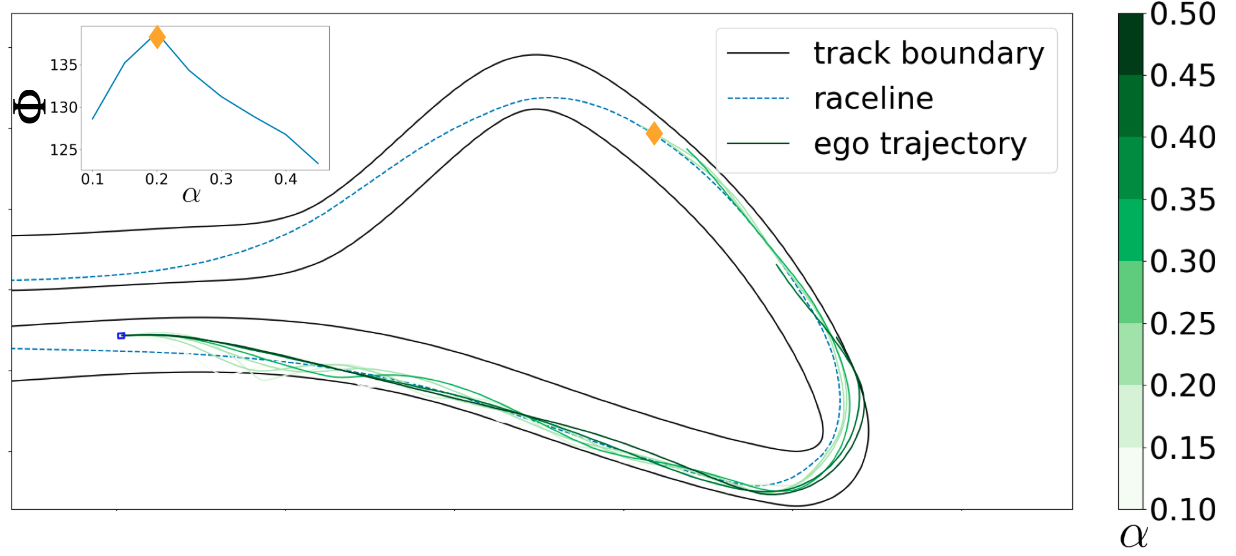}
        % {Figures/p_alpha_new.png}
        % \label{subfig:p_alpha}
    \end{subfigure}\\
    \begin{subfigure}{} 
        \includegraphics[width=5.5cm]{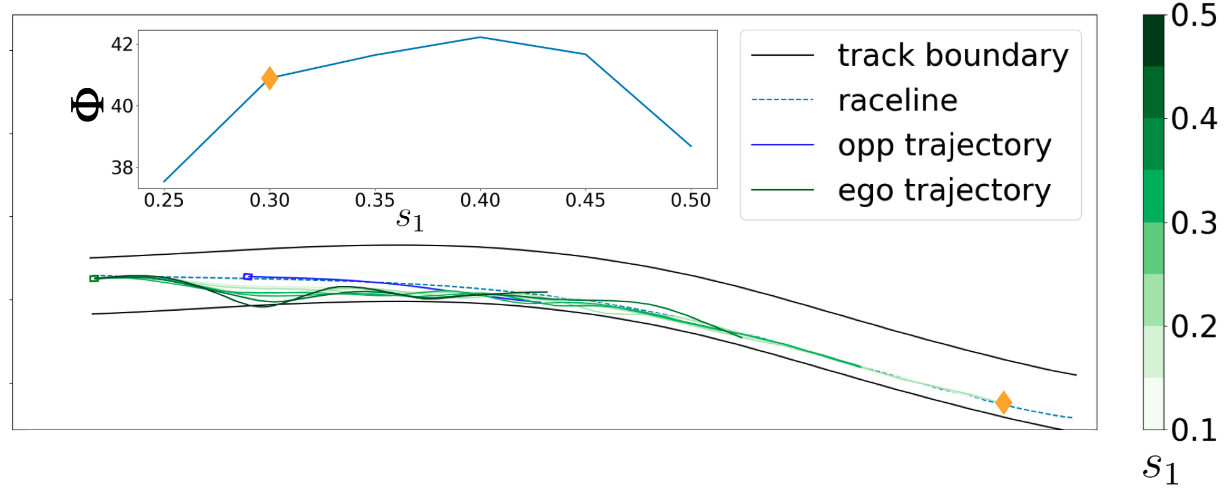}
        % {Figures/p_s1_new.png}
        % \label{subfig:p_s1}
    \end{subfigure}
    \begin{subfigure}{} 
        \includegraphics[width=5.5cm]{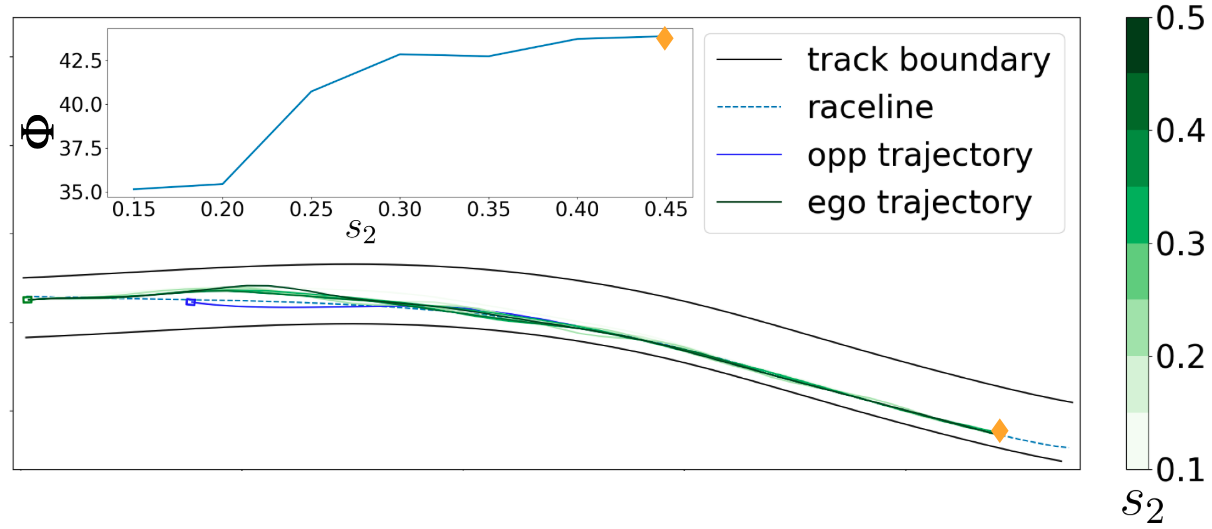}
        % {Figures/p_s2_new.png}
        % \label{subfig:p_s2}
    \end{subfigure}
    \begin{subfigure}{} 
        \includegraphics[width=8cm]{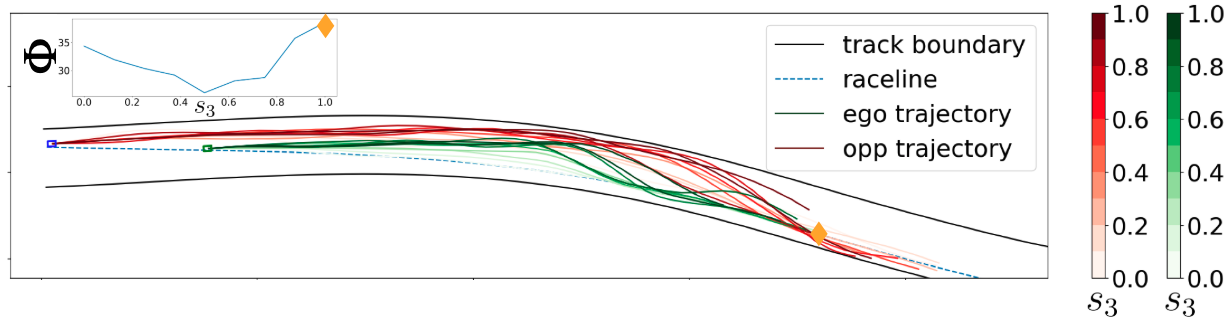}
        % \label{subfig:p_s3}
    \end{subfigure}
    \caption{Potential values and the trajectories at a given joint state for different (a) $q$ (b) $\zeta$ (c) $s_1$ (d) $s_2$ (e) $s_3$ of only the ego agent. We only denote 2 players here (only 1 player for (a) and (b)) and the 3rd player is far away from this position to not affect any players. Additionally, for ease of readability, we only show the impact of variation in trajectory of other player in response to \textsf{ego} in (e) as such deviations are not significant in (c) and (d).
    % We see that the trajectory that makes most progress correspond to approximate maximizing parameter for potential function (see yellow diamond).
    }
    \label{fig:potential_values}
    \vspace{-6mm}
\end{figure}

Next, in Figure~\ref{fig:potential_values}, we show that, by fixing all policy parameters but one, the parameter that (approx.) maximizes potential function  
% potential function maximized over different parameters of the ego agent
generates a trajectory that maximizes progress along the track over the next 5 seconds. This is highlighted by yellow diamond in Figure \ref{fig:potential_values}.

% \begin{figure}
%     \centering
%     \includegraphics[width=0.5\linewidth]{Figures/track.png}
%     \caption{Regrets histogram for a race. Maximum value of}
%     \label{fig:regrets}
% \end{figure}

\vspace{-5mm}
\subsection{Performance Comparison}
\vspace{-2mm}
To address questions \textbf{Q2} and \textbf{Q3}, we conduct 99 races involving three agents: \textsf{ego}, \(\textsf{O}_1\), and \(\textsf{O}_2\). Here, \textsf{ego} represents the agent using our proposed algorithm, while \(\textsf{O}_1\) and \(\textsf{O}_2\) are opponents employing other algorithms. In this study, we vary the opponents' algorithms and compute the winning fraction for the \textsf{ego} car. A visualization of the track used in the study, along with the computed optimal race-line, is provided in the Appendix (Figure~4).
The starting positions for the races are taken from three regions (denoted by \(\textsf{R}_1\), \(\textsf{R}_2\), and \(\textsf{R}_3\)), such that \(\textsf{R}_1\) is the furthest ahead on the track, followed by \(\textsf{R}_2\), and then \(\textsf{R}_3\), as shown in the Appendix (Figure~4). We perform 33 races, each with the \textsf{ego} agent starting in \(\textsf{R}_1\), \(\textsf{R}_2\), or \(\textsf{R}_3\). Also, \(\textsf{O}_1\) is always placed ahead of \(\textsf{O}_2\).

For the \textsf{ego}, we use a discount factor \(\gamma = 0.99\) and a training dataset comprising \(4000\) races. Below, we summarize five different opponent strategies for \(\textsf{O}_1\) and \(\textsf{O}_2\):

\noindent \((i)\) \textbf{Case I:} Opponents trained with a lower discount factor than \textsf{ego} (i.e., \(\gamma = 0.98\)); 

\noindent \((ii)\) \textbf{Case II:} Opponents trained with a higher discount factor than \textsf{ego} (i.e., \(\gamma = 0.995\));

\noindent \((iii)\) \textbf{Case III:} Opponents trained using fewer simulated races than \textsf{ego} (i.e., \(400\) races);

\noindent \((iv)\) \textbf{Case IV:} Opponents use the \textit{Iterated Best Response} (IBR) algorithm \footnote{The IBR algorithm used here is representative of the methods developed in \cite{spica2020real, Wang2020MultiagentSE}, though it may not be an exact implementation, as the original code is not publicly available.}, which computes the best response of opponents in round-robin for a fixed number of rounds (i.e., 6), with a planning horizon\footnote{Choice of hyperparameters is such that it roughly takes the same compute time as our approach.} of length \(2s\);

\noindent \((v)\) \textbf{Case V:} Opponents trained using self-play RL\footnote{Here, self-play RL represents the approach in \cite{thakkar2022hierarchical}, excluding the high-level tactical planner.}. We use a similar observation and reward as used in \cite{thakkar2022hierarchical} and train with \(2M\) steps.

\noindent The number of races won for all cases are provided in Table \ref{tab:3_agents}, where we see that \textsf{ego} agent trained using our approach has superior performance in comparison to all other opponent strategies.  

\vspace{1mm}
\noindent \textbf{Performance Variation with Hyperparameters (\textbf{Cases I–III}).}
The \textsf{ego} car outperforms opponents with lower discount factors (\textbf{Case I}) because lower discount factors lead to more myopic behavior, causing opponents to prioritize short-term progress over long-term track performance. Similarly, the \textsf{ego} car also outperforms opponents with higher discount factors (\textbf{Case II}) as higher discount factors increase the "effective horizon" of the game, which requires significantly more data to accurately approximate the value and potential functions.
Furthermore, the \textsf{ego} car has superior performance against opponents trained with less data of only $400$ races (\textbf{Case III}), as more data enables the learning of a more accurate \(\alpha\)-potential function.

\begin{table}[htbp] 
\centering
\begin{tabular}{|l|c|c|c|}
\hline
\textbf{Racing Scenario}              & \textbf{\# wins (ours)} & \textbf{\# wins ($\textsf{O}_1$)} & \textbf{\# wins ($\textsf{O}_2$)} \\ \hline
% {\color{red}MPC with random params}          & 78    & 12 &  9                   \\ \hline
% Opponents with low $\gamma$    
% \textbf{Case I} (Opponents with random policy parameters) & 84    & 9 & 6                   \\ \hline
\textbf{Case I} (Opponents with low $\gamma$) & 61    & 28 & 10                   \\ \hline
% Opponents with high $\gamma$  
\textbf{Case II} (Opponents with high $\gamma$)
& 52    & 40 & 7                   \\ \hline
% Opponents trained with less data    
\textbf{Case III} (Opponents trained with less data)
& 76    & 16 & 7  \\ 
\hline
% Opponent trained using IBR   
\textbf{Case IV} (Opponents using IBR)
& 73    & 22 & 4                   \\ \hline
% Opponent trained using self-play RL          
\textbf{Case V} (Opponents trained using self-play RL)
& 91    & 7 & 1                    \\ \hline
\end{tabular}
\caption{Outcomes of 99 races conducted between 3 cars under three different initial positions}
% :- a) ours $>$ $\textsf{O}_1$ $>$ $\textsf{O}_2$ b) $\textsf{O}_1$ $>$ ours $>$ $\textsf{O}_2$ c) $\textsf{O}_1$ $>$ $\textsf{O}_2$ $>$ ours, 33 races each. $\textsf{O}_1$ and $\textsf{O}_2$ are just 2 instances of the same opponent. 
% Videos of representative races in each scenario is at \href{https://sastry-group.github.io/alpha-RACE/}{\small https://sastry-group.github.io/alpha-RACE/}.}
\label{tab:3_agents}
\vspace{-3mm}
\end{table}

\begin{figure}
    \centering
    \includegraphics[width=.9\linewidth]{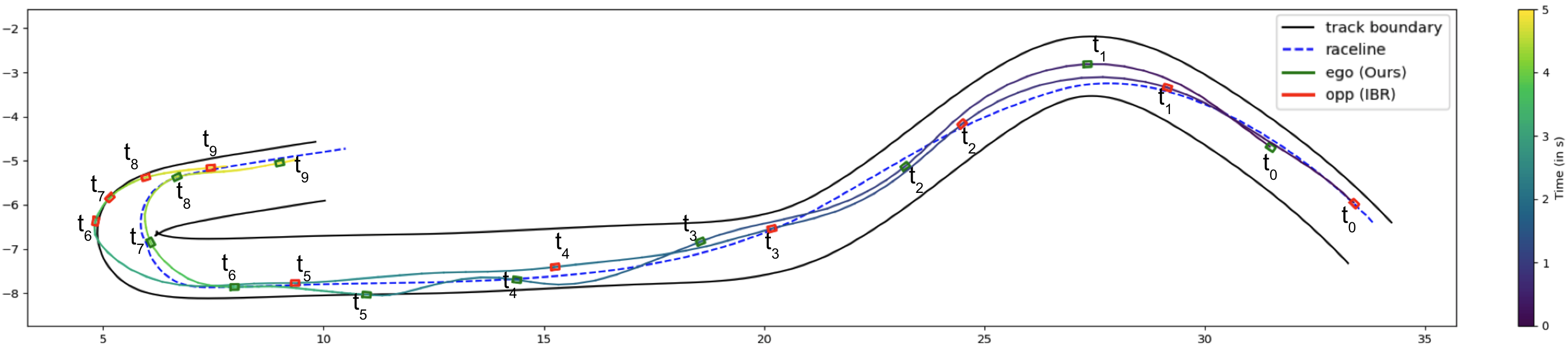}
    \caption{Example overtake in a race MPC vs IBR. Opponent (IBR) overtakes from $t_4$ to $t_5$ but later suffers at the turn from $t_5$ to $t_9$ when the Ego agent (Ours) overtakes back to re-claim it's position}
    \label{fig:race_eg}
    \vspace{-6mm}
\end{figure}

% \vspace{1mm}
\noindent \textbf{Comparison with opponents using IBR (Case IV).} Our approach surpasses IBR for two key reasons. First, IBR may not always converge to a Nash equilibrium. Second, IBR's real-time computation is often restricted to local planning with a short \(2s\) horizon, as noted in \cite{spica2020real, wang2019bgame}. In contrast, competitive racing demands prioritizing a global racing line to optimize long-term performance, with strategic deviations for overtaking or blocking rather than short-term gains. This distinction is highlighted in Figure~\ref{fig:race_eg}, where the IBR player achieves higher straight-line speeds to overtake but struggles in turns due to aggressive braking. While our implementation is not a direct comparison with \cite{spica2020real}, our approach supports longer-horizon planning for real-time control by leveraging offline potential function learning. This is achieved by constraining the policy space to primitive behaviors, closely aligning with practical racing strategies.

\noindent \textbf{Comparison with opponents using self-play RL (Case V).}
Our approach also outperforms opponents trained via self-play RL. While we do not claim an optimized implementation of self-play RL—acknowledging potential improvements through further training, hyperparameter tuning, or tactical enhancements as in \cite{thakkar2022hierarchical}—our method offers clear advantages. It delivers interpretable solutions grounded in realistic racing strategies and approximation guarantees to a Nash equilibrium in races with more than 2 cars. In contrast, self-play RL often requires additional insights to develop effective policies. For example, \cite{thakkar2022hierarchical} showed that augmenting self-play RL with high-level tactical plans significantly enhances policy learning.

% \acks{We thank }

\vspace{-5mm}
\section{Conclusion}\label{sec: Conclusion}
\vspace{-2mm}
In this work, we study real-time algorithms for competitive multi-car autonomous racing. Our approach is built on two key contributions: first, a novel policy parametrization and utility function that effectively capture competitive racing behavior; second, the use of dynamic \(\alpha\)-potential functions to develop real-time algorithms that approximate Nash equilibrium strategies. This framework enables the learning of equilibrium strategies over long horizons at different game states through the maximization of a potential function.

There are several directions of future research. One interesting direction is to integrate the strengths of self-play RL discussed in \cite{thakkar2022hierarchical}, with our approach. For instance, potential function maximization could guide self-play RL as a high-level reference plan generator. Alternatively, our method could initialize self-play policies via behavior cloning or refined reward design, such as using the potential function's maximum value as a reward signal or leveraging GAIL \cite{Ho2016GenerativeAI} with expert trajectories derived from our approach.
\bibliography{refs}

\newpage

\appendix
\section{Proof of Proposition \ref{prop: NE_def_PotMax}}
    Consider a policy parameter \(\theta^\ast\) such that \(\Phi(\mathbf{x},\theta^\ast) \geq \max_{\theta\in \Theta}\Phi(\mathbf{x}, \theta) - \lambda \). For any \(\theta^i\in \Theta^i\),
    \begin{align*}
        V^i(\mathbf{x},\theta^{\ast,i},\theta^{\ast, -i}) - V^i(\mathbf{x},\theta^i,\theta^{\ast, -i})& \geq  \Phi(\mathbf{x},\theta^{\ast,i},\theta^{\ast, -i}) - \Phi(\mathbf{x},\theta^i,\theta^{\ast, -i}) - \alpha \\ 
        &\geq  -\lambda - \alpha,
    \end{align*}
    where first inequality is due to \eqref{eq: NearMPGDef} and the second inequality is because \(\theta^\ast\) maximizes \(\Phi\). The proof follows using the definition of Nash equilibrium (Definition \ref{def: NE}).

\section{Description of Single-agent Racing Line}\label{sec:racing_line}

Race drivers follow a racing line for specific maneuvers. This line can be used as a reference path by the motion planner to assign time-optimal trajectories while avoiding collision. The racing line is minimum-time or minimum-curvature. They are similar, but the minimum-curvature path additionally allows the highest cornering speeds given the maximum legitimate lateral acceleration \cite{doi:10.1080/00423114.2019.1631455}. 

There are many proposed solutions to finding the optimal racing line, including nonlinear optimization \cite{Rosolia2020LearningHT,doi:10.1080/00423114.2019.1631455}, genetic algorithm-based search \cite{Vesel2015RacingLO} and Bayesian optimization \cite{Jain2020ComputingTR}. However, for our work, we calculate the minimum-curvature optimal line, which is close to the optimal racing line as proposed by \cite{doi:10.1080/00423114.2019.1631455}. 
%The race track information is input by a sequence of tuples ($x_i$,$y_i$,$w_i$), $i \in \{0,...,N-1\}$, where ($x_i$,$y_i$) denotes the coordinate of the center location and $w_i$ denotes the lane width at the $i^{th}$ point, vehicle width $w_{veh}$. The output trajectory consists of a tuple of seven variables: coordinates $x$ and $y$, curvilinear longitudinal displacement $s$, longitudinal velocity $v_x$, acceleration $a_x$, heading angle $\psi$, and curvature $\kappa$. The trajectory is obtained by minimizing the following cost: In our work, we calculate the minimum-curvature optimal line, similar to \cite{doi:10.1080/00423114.2019.1631455}. 
The race track is represented by a sequence of tuples ($x_i$,$y_i$,$w_i$), $i \in \{0,...,N-1\}$, where ($x_i$,$y_i$) denotes the coordinate of the center location and $w_i$ denotes the lane width at the $i$-th point. The output racing line consists of a tuple of seven variables: coordinates $x$ and $y$, longitudinal displacement $s$, longitudinal velocity $v_x$, acceleration $a_x$, heading angle $\psi$, and curvature $\kappa$. It is obtained by minimizing the following cost:
%\vspace{-0.2cm}
\begin{equation} \label{opt_racing_line_eqn1}
\begin{split}%\label{total cost term}
    &  \mathop{\min}\limits_{\eta_1...\eta_N} \quad \sum_{n=0}^{N-1} \kappa_i^2(n)\\
    \text{s.t.} & \quad \eta_i \in \left[ -\frac{w_i}{2}+\frac{w_{veh}}{2},\frac{w_i}{2}-\frac{w_{veh}}{2} \right]\\
\end{split}
\end{equation}
where the vehicle width is $w_{veh}$, and $\eta_i$ is the lateral displacement with respect to the reference center line.

To create a velocity profile, we need to consider the vehicle's constraints on both longitudinal and lateral acceleration \cite{doi:10.1080/00423114.2019.1631455}. Our approach involves generating a library of velocity profiles, each tailored to specific lateral acceleration limits determined by the friction coefficients for the front ($\mu_f$) and rear ($\mu_r$) tires, as well as the vehicle's mass ($m$) and the gravitational constant ($g$). In particular, we produce a set of velocity profiles covering a range of maximum lateral forces corresponding to the friction $\mu_{eff}$ within the interval $[\mu_{min}, \mu_{max}]$. This library allows us to retrieve a velocity profile that matches a given value of $\mu$. Interpolation is necessary when we encounter a friction value that falls within the valid range but is not explicitly present in the library.

\begin{figure}
    \centering
    \includegraphics[width=\linewidth]{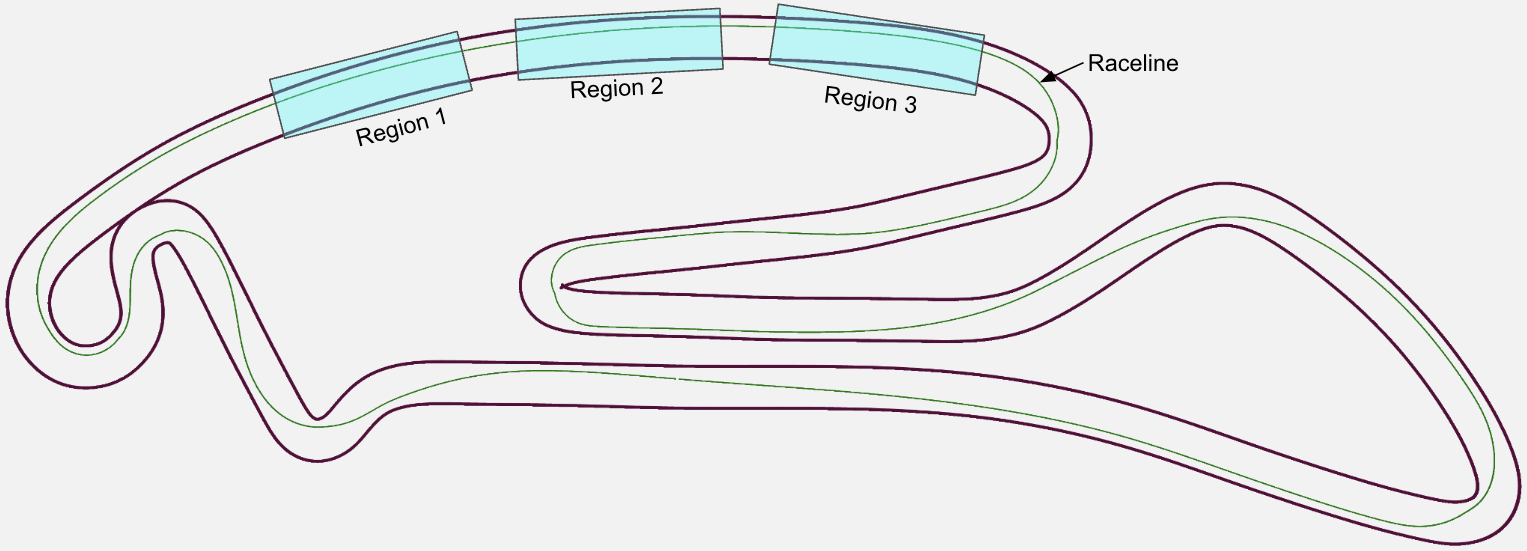}
    \caption{Track and the starting position regions}
    \label{fig:track}
\end{figure}

An example of a racing line calculated for the racetrack used in our numerical study in Section \ref{sec: Numerical} is shown in Figure \ref{fig:track}. 

\section{Dynamic Bicycle Model}
For any car \(i\), we denote its mass by \(m^i\), its moment of inertia in the vertical direction about the center of mass by \(I_z^i\), the distance between the center of mass (COM) and its front wheel by \(l_f^i\), and the distance from the COM to the rear wheel \(l_r^i\). Also, $\kappa_t^i$ denotes the inverse of radius of curvature of the track at $p_{x,t}^i$. Using these notations, the dynamics of car \(i\) is defined below:
\begin{equation} \label{eqn:dyn_model}
\begin{split}
&\begin{bmatrix}
    {p}_{x,t+1}^i \\
    {p}_{y,t+1}^i \\
    \phi_{t+1}^i \\
    \tilde{v}_{x,t+1}^i \\
    \tilde{v}_{y,t+1}^i \\
    \omega^i_{t+1}
\end{bmatrix}
= \begin{bmatrix}
    {p}_{x,t}^i \\
    {p}_{y,t}^i \\
    \phi_{t}^i \\
    v_{x,t}^i \\
    v_{y,t}^i \\
    \omega^i_{t}
\end{bmatrix} + \Delta t
\begin{bmatrix}
    v_{x,t}^i\\
    v_{y,t}^i\\
    \omega_t^i - \frac{\kappa_{t}^i}{(1-\kappa_{t}^i p_{y,t}^i)}(\tilde{v}_{x,t}^i \cos (\phi_t^i) - \tilde{v}_{y,t}^i \sin (\phi_t^i))\\
    \frac{1}{m^i} (F_{r,x,t}^i - F_{f,y,t}^i \sin(\delta_t^i) + m^i \tilde{v}_{y,t}^i \omega_t^i)\\
    \frac{1}{m^i} (F_{r,y,t}^i + F_{f,y,t}^i \cos(\delta_t^i) - m^i \tilde{v}_{x,t}^i \omega_t^i)\\
    \frac{1}{I_z^i} (F_{f,y,t}^i l_f^i \cos(\delta_t^i) - F_{r,y,t}^i l_r^i) 
\end{bmatrix} \\ 
\end{split},
\end{equation}
where 
\((i)\) \(v_{x,t}^i = \frac{1}{(1-\kappa_{t}^i p_{y,t}^i)}(\tilde{v}_{x,t}^i \cos (\phi_t^i) - \tilde{v}_{y,t}^i \sin (\phi_t^i))\), \(v_{y,t}^i = \tilde{v}_{x,t}^i \sin (\phi_t^i) + \tilde{v}_{y,t}^i \cos (\phi_t^i)\) are the velocities in frenet frame; \((ii)\) \(\tilde{v}_{x,t}^i, \tilde{v}_{y,t}^i\) are velocities in body frame; \((iii)\) \(
F_{r,x,t}^i = (C_{1} - C_{2} \tilde{v}_{x,t}^i) d_t^i - {C_{3}} - {C_4} (\tilde{v}_{x,t}^i)^2
\)
is the longitudinal force on the rear tire at time \( t \). Here, \( C_1 \) and \( C_2 \) are parameters that govern the longitudinal force generated on the car in response to the throttle command, while \( C_3 \) and \( C_4 \) are parameters that account for the friction and drag forces acting on the car; \((iv)\) $F_{f,y,t}^i = D_f \sin(C_f \tan^{-1}(B_f \alpha_{f,t}^i))$ is the lateral force on the front tire depending on the slipping angle \(\alpha_{f,t}^i\), which is given by $\alpha_{f,t}^i = \delta^i_t - \tan^{-1}\left(\frac{\omega_t^i l_f + \tilde{v}_{y,t}^i}{\tilde{v}_{x,t}^i}\right)$. Here $B_{f}, C_f, D_f$ are the parameters of Pacejka tire model; and \((v)\)
$F_{r,y,t}^i = D_r \sin(C_r \tan^{-1}(B_r \alpha_{r,t}^i))$ is the lateral force on the rear tire depending on the slipping angle \(\alpha_{r,t}^i\), which is given by $\alpha_{r,t}^i = \tan^{-1}\left(\frac{\omega_t^i l_r - \tilde{v}_{y,t}^i}{\tilde{v}_{x,t}^i}\right)$. Here $B_r, C_r, D_r$ are the parameters of Pacejka tire model.

% \todo[inline]{
% 1. Experimental details in appendix 

% 2. Add about appendix at relevant places 

% 3. Algorithms for self-play RL and IBR 

% }

\section{Hyperparameters}
\subsection{Network architecture}

We use a simple feed-forward deep neural network with ReLU activation except for the last layer to represent the value function and the potential function. The network for value function consists of $3$ hidden layers with $(128,128,64)$ hidden features on each layer. The network for potential function consists of $3$ hidden layers with $(384,384,192)$ hidden features on each layer. 

\subsection{Training}

We use a learning rate of $0.0001$ and train for $50000$ epochs (both value functions and potential function). Each race consists of $500$ time steps with $dt=0.1s$, hence $50s$ race.

\section{Self-play RL training}

We use standard PPO training parameters as available in stable\_baselines3 with batch size $1024$, number of epochs $5$, learning rate $0.0005$, $\gamma=0.99$ and $8$ environments in parallel. The observation used is the same as the joint state input used for our work for fair comparison. The reward design used is also the same as the utility used in our work. We train for $100K$ time-steps for each iteration of self-play RL where we switch agents for training for total of $99$ times i.e. $33$ cycles of training for $3$ agents

\section{Iterated Best Response (IBR) hyperparameters}

We use $N=6$ iterations for iterated best response with the same utility as the one used in our work and with horizon length of $2s$ with $20$ time-steps of length $dt=0.1s$. The solve time with the following parameters is $0.1s$ which is comparable to the compute time required by our algorithm

\section{Table of Notations}
\begin{longtable}{|l|l|}
\hline
\textbf{Notation} & \textbf{Description} \\
\hline
\endfirsthead
\hline
\textbf{Notation} & \textbf{Description} \\
\hline
\endhead
\hline
\endfoot
\hline
\endlastfoot

\(\mathbf{x}^i\) & State vector of car \(i\), including position, velocity, and angular velocity. \\
\(p_x^i, p_y^i\) & Longitudinal and lateral positions of car \(i\) in the global frame. \\
\(\phi^i\) & Orientation of car \(i\) in the global frame. \\
\(v_x^i, v_y^i\) & Longitudinal and lateral velocities of car \(i\) in the body frame. \\
\(\omega^i\) & Angular velocity of car \(i\) in the global frame. \\
\(\mathbf{u}^i = (d^i, \delta^i)\) & Control input for car \(i\), where \(d^i\) is the throttle and \(\delta^i\) is the steering angle. \\
\(d^i\) & Throttle input of car \(i\). \\
\(\delta^i\) & Steering angle of car \(i\). \\
\(d_{\min}, d_{\max}\) & Minimum and maximum throttle limits for car \(i\). \\
\(\delta_{\min}, \delta_{\max}\) & Minimum and maximum steering angle limits for car \(i\). \\
\(m^i\) & Mass of car \(i\). \\
\(I_z^i\) & Moment of inertia of car \(i\) in the vertical direction about the center of mass. \\
\(l_f^i\) & Distance from the center of mass to the front wheel of car \(i\). \\
\(l_r^i\) & Distance from the center of mass to the rear wheel of car \(i\). \\
\(F_{r,x,t}^i\) & Force applied to the rear wheel of car \(i\) in the longitudinal direction at time \(t\). \\
\(F_{r,y,t}^i\) & Force applied to the rear wheel of car \(i\) in the lateral direction at time \(t\). \\
\(F_{f,y,t}^i\) & Force applied to the front wheel of car \(i\) in the lateral direction at time \(t\). \\
\(\Delta t\) & Time step for the simulation. \\
\(\gamma\) & Discount factor in the utility function. \\
\(\mathbf{x}_t\) & State of the system at time \(t\). \\
\(\textbf{u}_t^i\) & Control input of car \(i\) at time \(t\). \\
\(\omega_t^i\) & Angular velocity of car \(i\) at time \(t\). \\
\(\theta^i\) & Parameter of car \(i\)'s policy. \\
\(\Theta^i\) & Set of possible parameters for car \(i\)'s policy. \\
\(\Pi^i\) & Set of possible strategies for car \(i\). \\
\(\Pi\) & Set of joint strategies for all players (cars). \\
\(\epsilon\) & The tolerance used in the definition of \(\epsilon\)-Nash equilibrium. \\
\(V^i(\mathbf{x}, \theta^i, \theta^{-i})\) & The expected long-run utility of car \(i\) given the state \(\mathbf{x}\) and strategy profile \(\theta\). \\
\(\theta^\ast\) & The \(\epsilon\)-Nash equilibrium strategy profile. \\
\end{longtable}
\end{document}